\documentclass[preprint,12pt,authoryear]{elsarticle}

\usepackage{amssymb}
\usepackage{array}

\usepackage{url}

\journal{Safety Science}

\begin{document}

\begin{frontmatter}

\title{Hazard analysis of human-robot interactions with HAZOP-UML}

\author{J\'er\'emie Guiochet}

\address{University of Toulouse, LAAS-CNRS, Toulouse, France}
 \ead{jeremie.guiochet@laas.fr}
 \ead[url]{homepages.laas.fr/guiochet}

\begin{abstract}
New safety critical systems are about to appear in our everyday life: advanced robots able to interact with humans and perform tasks at home, in hospitals, or at work. A hazardous behavior of those systems, induced by failures or extreme environment conditions, may lead to catastrophic consequences. Well-known risk analysis methods used in other critical domains (e.g., avionics, nuclear, medical, transportation), have to be extended or adapted due to the non-deterministic behavior of those systems, evolving in unstructured environments. One major challenge is thus to develop methods that can be applied at the very beginning of the development process, to identify hazards induced by robot tasks and their interactions with humans. In this paper we present a method which is based on an adaptation of a hazard identification technique, HAZOP (Hazard Operability), coupled with a system description notation, UML (Unified Modeling Language). This systematic approach has been applied successfully in research projects, and is now applied by robot manufacturers. Some results of those studies are presented and discussed to explain the benefits and limits of our method.

\end{abstract}

\end{frontmatter}

\section{Introduction}

\label{sec:intro}
Besides the developments of well-known safety critical systems in aeronautics or transportation, new systems are about to appear in our everyday life: robots at home, at work, or in the hospitals \citep{ROY15}. 
Such systems, will interact with users, and execute tasks in the vicinity or even in physical contact with humans. Hence, a failure of such complex systems may lead to catastrophic consequences for users which is a major obstacle to their deployment in real life. Most safety analysis techniques coming from the dependability \citep{AVI04} or risk management \citep{ISO31000} domains could be used for such systems, but some specificities of robots limit their efficiency. For instance, the fact that robots move in unstructured and unknown environments makes the verification and validation (mainly through testing) non sufficient (it is impossible to guarantee that all main scenarios have been tested); the presence of users and complex non deterministic software (with decisional mechanisms) limit the use of quantitative risk analysis techniques; classical hazard analysis techniques are also not adapted to the complexity of human-robot interactions. Little work has been done about risk analysis for such systems, although it is a major challenge for robot certification \citep{MIT12}. Many robotics studies about estimation and treatment of collision risks exist (many references presented by \cite{HAD14}), but few are on risk analysis methods \citep{DOG14}. The safety community has rarely addressed this issue, whereas we have been working on this for a decade \citep{GUIIARP02,GUIHESSD04}.

Some robot manufacturers use directives \citep{2006/42/EC} or standards \citep{ISO13849} dedicated to machines, but they are not completely applicable, particularly when there is a human-robot physical interaction. Generic standards like \cite{IEC61508b}, are also hardly applicable due to uncertainties in the robot behavior (in this standard, fault correction through artificial intelligence is not recommended for safety integrity level SIL2 to SIL4). More recently, the  standard \cite{ISO10218} for industrial robots that might share their workspace with humans, has been completed by the \cite{ISO13482}. It is also important to note that such standards, do not cover other application domain robots. For instance, in the medical field, there is no robotic-specific standard, and the robots are considered as active medical devices such as defined in the \cite{93/42/CEEang}, and covered by \cite{14971ang2006} for risk management. In all those standards, classic risk management and design recommendations are proposed, but no specific guidelines for risk analysis techniques are presented. 

To cope with the previous issues, we suggest a hazard identification technique with the following objectives:
\begin{enumerate}
\item applicable from the very beginning of the development process
\item includes human activity as a source of hazard 
\item provides guidance for analysts with list of guide words 
\item focuses on operational hazards, i.e., hazards linked with the robot tasks and interactions 
\end{enumerate}
Among risk analysis techniques, the most widely used are Preliminary Hazard Analysis (PHA), Hazard Operability Analysis (HAZOP), Fault Tree Analysis (FTA), and Failure Mode, Effects, and Criticality Analysis (FMECA). The two first may be applied as hazard analysis at the very early steps of a development process, whereas FTA and FMECA are more dedicated to advanced steps, focusing more on reliability aspects. Thus, we chose to base our method on HAZOP, and to combine it with the system modeling language UML (Unified Modeling Language). This method developed at LAAS \citep{GUI10, MAR10, GUI13}, has been successfully applied in several French and European projects \citep{PHRIENDS,SAPHARI,MIRAS} in collaboration with robot manufacturers  (KUKA Robotics, AIRBUS Group and Robosoft). This paper synthesizes for the first time our work on HAZOP-UML, and proposes an analysis of the applications in these projects. 

The remainder of this paper is structured as follows. Section~\ref{sec:background} provides background on UML  and HAZOP. In Section~\ref{sec:hazopuml}, we present the HAZOP-UML method, and in Section~\ref{sec:exp}, results of several experiments are analyzed and discussed. In Section~\ref{sec:relatedwork}, related work on model-based safety analysis is compared to our approach. We conclude in Section~\ref{sec:conclusion} by outlining the benefits and limits of HAZOP-UML, and listing some future directions.

\section{Background}
\label{sec:background}

\subsection{Unified Modeling Language}
\label{sec:UML}
UML (Unified Modeling Language) is a graphical notation, widely used in software and system engineering domains to support early steps of the development process. Its specification is available on the Object Management Group UML page\footnote{www.uml.org : accessed 2015-05-15}. The current version (UML 2), has thirteen diagrams, that could be classified in static diagrams (e.g., class diagram) and dynamic diagrams (e.g., use case, sequence and state machine diagrams). UML is a language, and not a method, as it is not specified in which chronological order each diagram must be used. But, use cases and sequence diagrams are typically used at the beginning of any project development. State machine diagrams are also widely used in reactive systems as robot controllers. Hence, we will present those three diagrams, focusing only in the elements we will use for our approach. One main pitfall using this language is to mix different levels of details in the same diagram. For instance, mixing some high level specifications with  implementation constraints on the same diagram is error prone and also not recommended for the safety analysis. This is why we also put forward in this paper some modeling rules to avoid this pitfall and to guide the analysts.  

\begin{figure*}
	\center
	\includegraphics[width=4cm]{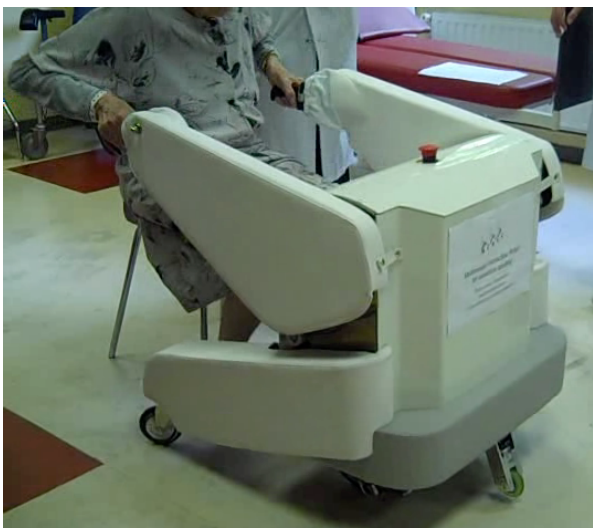}
	\caption{MIRAS robot prototype during clinical investigation}
	\label{fig:miras-robot}
\end{figure*}

As a running example, we will use some models of the case study \cite{MIRAS}, an assistive robot presented Figure~\ref{fig:miras-robot}, for
standing up, sitting down and walking, and also capable of health-state monitoring of the patients. It
is designed to be used in elderly care centers by people suffering
from gait and orientation problems where a classic wheeled walker (or ``rollator"), is not sufficient for patient autonomy. The robotic rollator is composed of a mobile base
and a moving handlebar. 

\begin{figure}
\centering
\includegraphics[width=8cm]{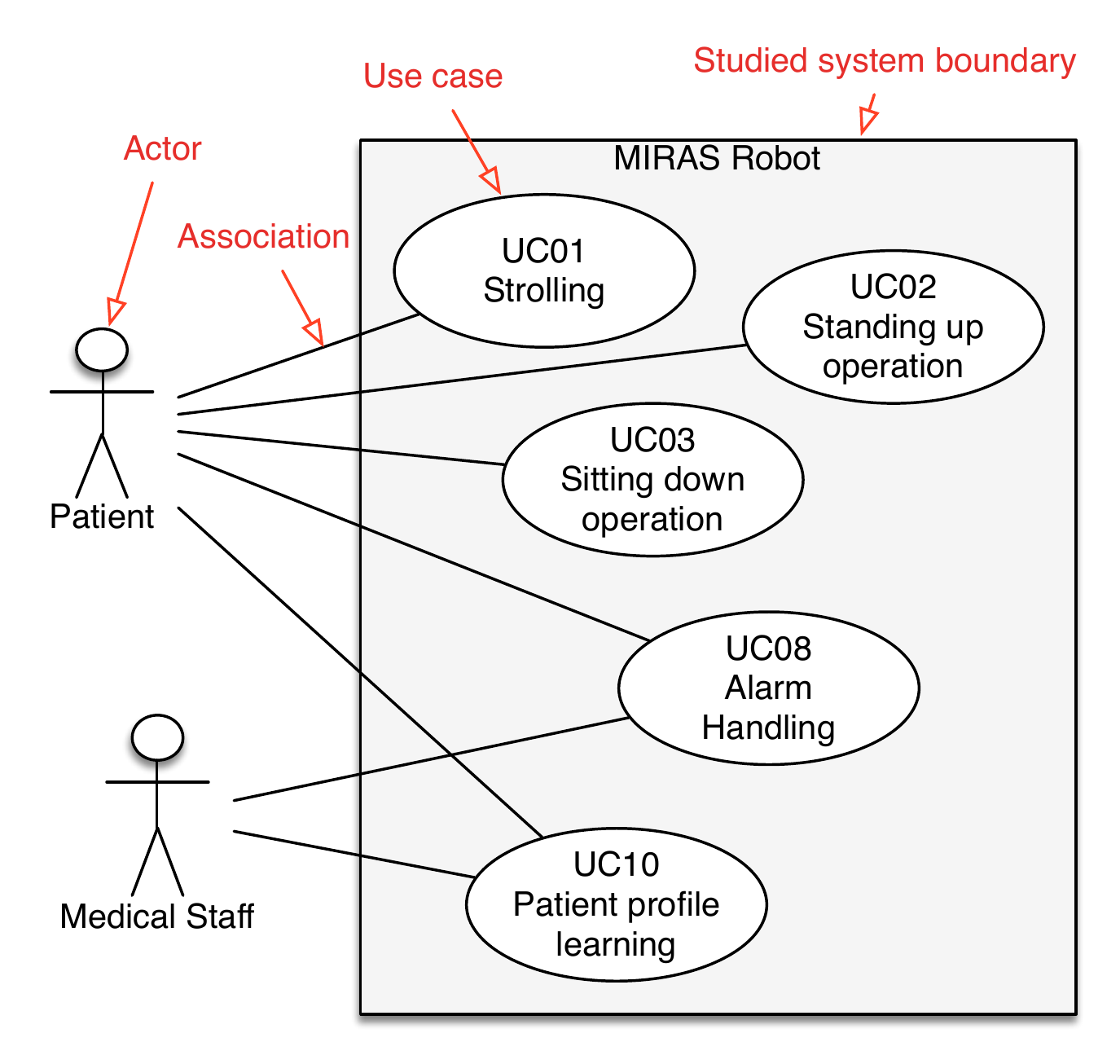}
\caption{Extract of MIRAS use case diagram from \cite{GUI13}}
\label{fig:ucdiag}
\end{figure}

\paragraph{Use case diagrams}
This diagram is the basic requirement UML model, presenting the system to analyse, the actors communicating with it, and the objectives for the use of the system: the use cases. The example of Figure~\ref{fig:ucdiag} only presents a subset of the complete use case diagram (15 use cases), and the two involved actors. In this diagram, the proposed services are to help the patient to stand up (UC02), deambulate (UC01), and sit down (UC03). The system is also able to detect physiological issues and trigger an alarm (patient heartbeat and fatigue, in UC08). We also represent that the system offers the profile learning facility (UC10). 
In some projects using UML the mechanical part of a robot is represented as a UML actor, and the system boundary (the box around use cases) defines the robot controller (including software and hardware). We do not recommend using such an approach to perform the hazard identification, indeed, the complete system has to be studied as a whole. 

\begin{figure}
\centering
\includegraphics[width=12cm]{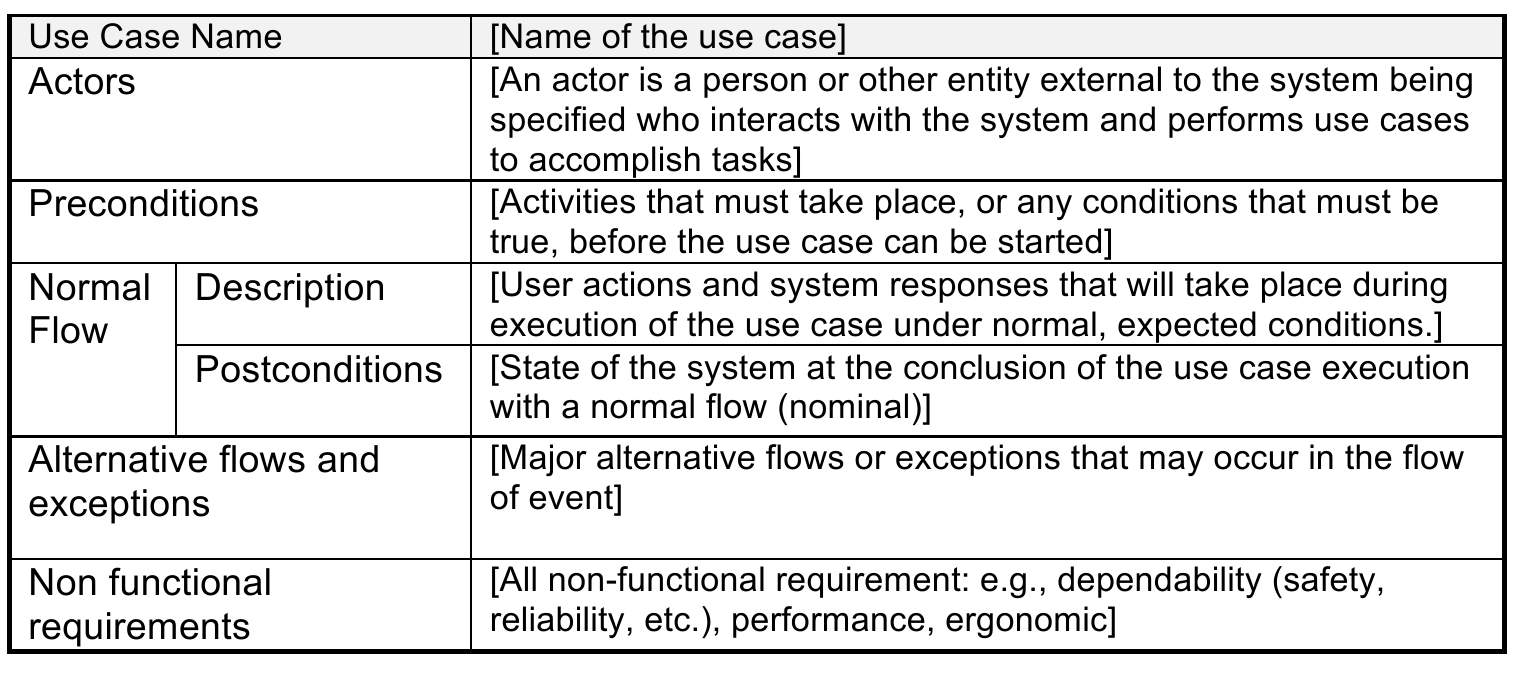}
\caption{Use case textual description template}
\label{fig:ucfile}
\end{figure}

This diagram provides an expressive and simple mean to communicate between developers, analysts and users. This graphical representation is always completed with a textual description as in Figure~\ref{fig:ucfile}. Important information such pre and post conditions, and non-functional requirements are included. Use case diagram only represents functional requirements. Textual description of the normal, alternative and exception flows may also be presented with sequence diagrams as presented hereafter. 

\begin{figure}
\centering
\includegraphics[width=8cm]{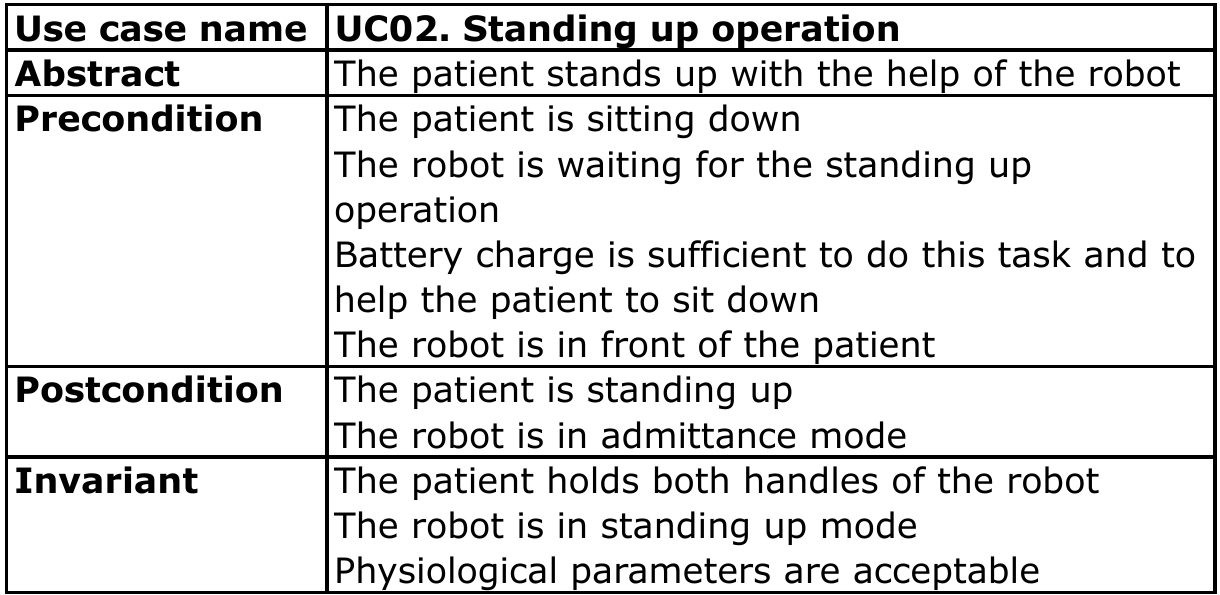}
\caption{UC02 use case textual description with pre,post conditions and invariant}
\label{fig:ucfilehazop}
\end{figure}

In the UML OMG standard, some relations may exist between use cases (mainly the relations \emph{extend} and \emph{include}) but we recommend not to use them, as they often lead to misunderstandings and to an unclear application of the HAZOP-UML method. In order to prepare the HAZOP-UML study, an extract from the use case textual description should be done, with only the pre and post conditions, and also the invariants coming from safety properties in the ``Non functional requirements" category. An example of such a table is given in Figure~\ref{fig:ucfilehazop} for the UC02 of the MIRAS running example.


\begin{figure}
\centering
\includegraphics[width=10cm]{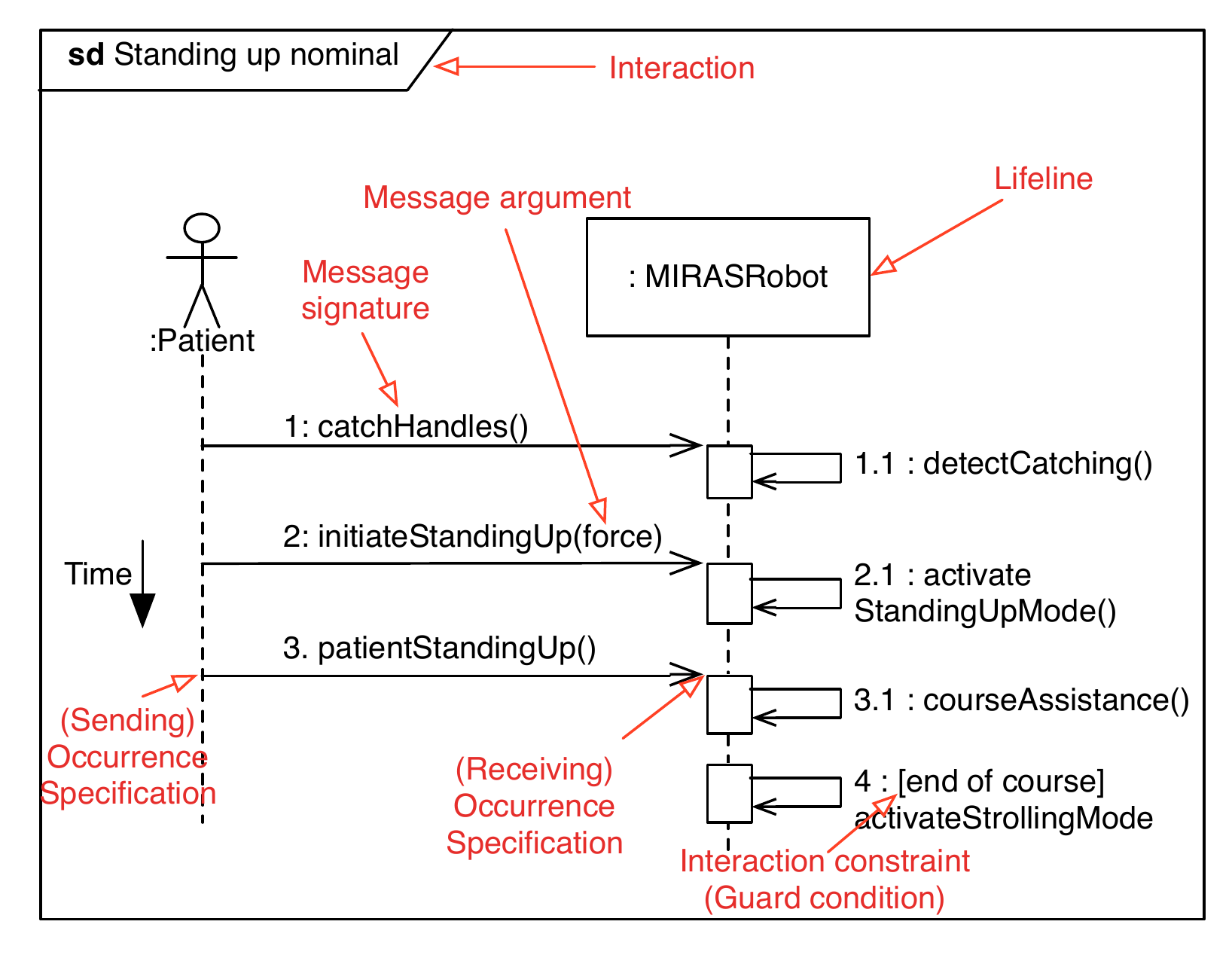}
\caption{Sequence diagram for the nominal scenario of \emph{UC01: Standing up operation}}
\label{fig:sequencediag}
\end{figure}

\paragraph{Sequence diagrams} 
Figure~\ref{fig:sequencediag} shows a sequence diagram, describing a possible scenario, which is actually an instance of an UML use case. This diagram shows a nominal scenario for the UC02. Other scenarios are possible for the UC02, like alternative flow of events (e.g., the patient releases the handles while she is standing up). This second scenario will be represented with another sequence diagram (not presented here). The expressiveness of such diagram is well adapted to represent human-robot interactions, and have proven to be useful while discussing with other stakeholders who are not experts in this language (doctors, mechanical engineers, etc.). All messages exchanged between actors and the system are represented along their lifelines. In our case three types of messages are used:
\begin{itemize}
\item \textit{indirect interaction} through robot teach pendant (hardware or software interfaces)
\item \textit{cognitive interaction}, e.g., gesture or voice/audio signals are exchanged
\item \textit{physical interaction}, direct contact between physical structure of the robot and the user
\end{itemize}
In the example of Figure~\ref{fig:sequencediag}, the messages are all physical contacts, so we did not add this information which can be done using a UML annotation. 
In UML, a sequence diagram is a representation of an \emph{Interaction}, where actors and the system (\emph{Lifeline}), send some \emph{Message} that might have \emph{Arguments} and \emph{Constraints}. Here the message \emph{2:initiateStandingUp} is sent to the robot with a \emph{force} exerced on the handles. As the time increases from top to bottom, each message has a \emph{sending} and \emph{receiving occurrence event}. It is also possible to represent on a message a \emph{guard condition} for its execution (e.g., \emph{[end of course]} of message 4). 

\begin{figure}
\centering
\includegraphics[width=11cm]{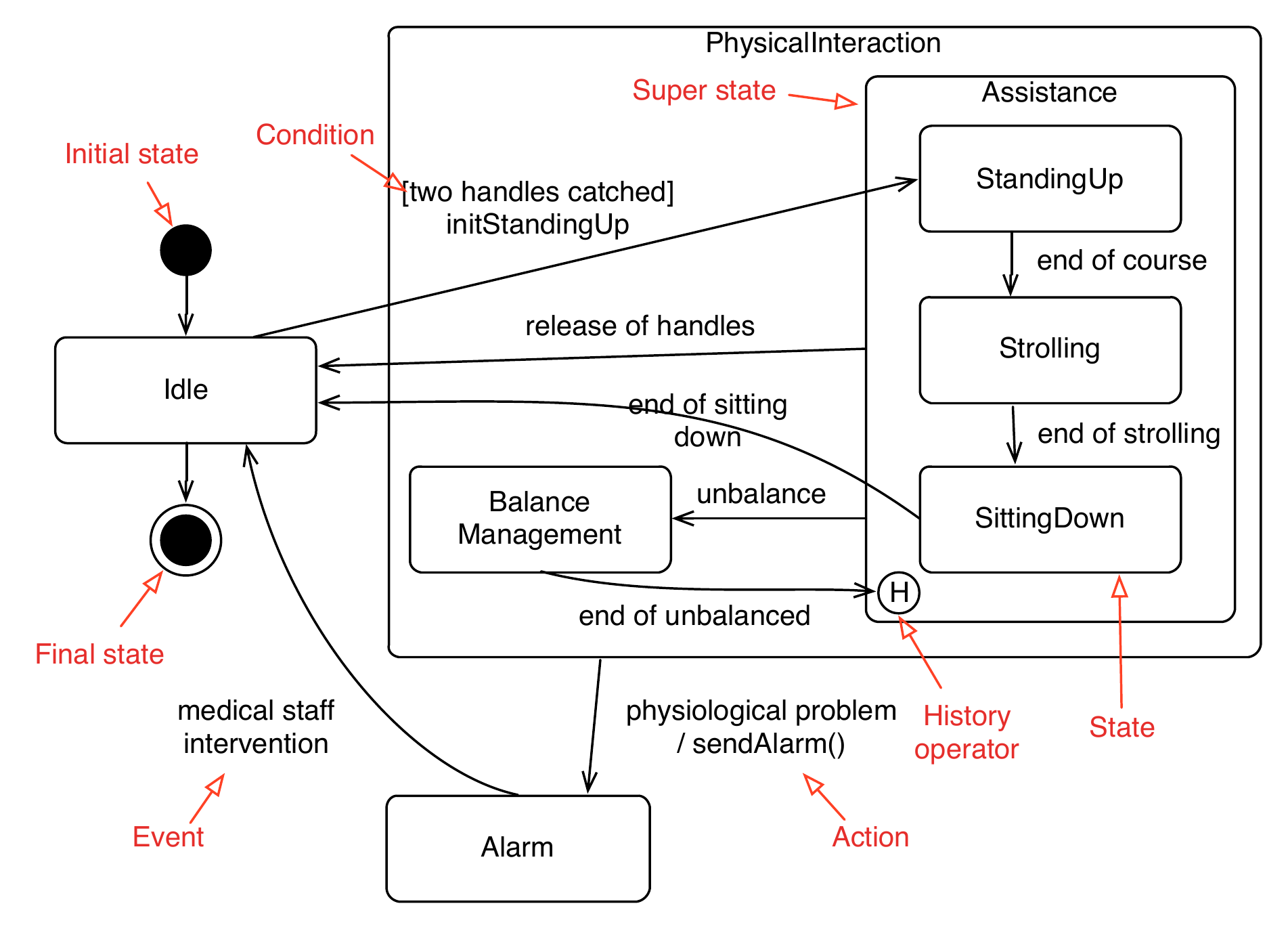}
\caption{Simplified version of MIRAS state machine}
\label{fig:statemachine}
\end{figure}

We recommend not to use the UML2 \emph{fragments} (loops, alternatives, etc.) but to rather use several diagrams to represent alternatives flows for instance. We also recommend to draw a \emph{system sequence diagram}, i.e., representing only the actors and the system, and not the internal objects of the system.
\paragraph{State machines} 
These deterministic automata diagrams are based on the statecharts proposed by \cite{HAREL}. A state machine is given for all the objects with a dynamic behavior. An example is given in Figure~\ref{fig:statemachine} where the considered object is the MIRAS robot controller. 
A transition is represented with an arrow between a start state and a destination state, and can have the following facultative form of {\tt event [guard] / action()}, where:
		\begin{itemize}
		\item {\tt event} is the trigger element of the transition, which could be:
			\begin{itemize}
				\item \textit{signal event}: asynchronous external event  (e.g., button pressed, voice command)
				\item \textit{call event}: reception of an operation called by another object of the system
				\item \textit{change event}: a change of a boolean variable based on the estimation of a system variable
				\item temporal event (\textit{after} or \textit{when}): expired duration \emph{after($<$duration$>$)}, or absolute time \emph{when(date=$<$date$>$)}
			\end{itemize}
		\item {\tt guard} is a condition estimated only if the event occurs
		\item {\tt action} is a list of actions performed instantly when the transition is triggered
		\end{itemize}

In this method we use state diagrams to specify at the beginning of a project, the different operational modes of the robot. This diagram is also useful for the detailed design and implementation of the robot controller, which is out of the scope of this paper.

\subsection{HAZOP}
\label{sec:hazop}

\begin{figure}
\centering
\includegraphics[width=10cm]{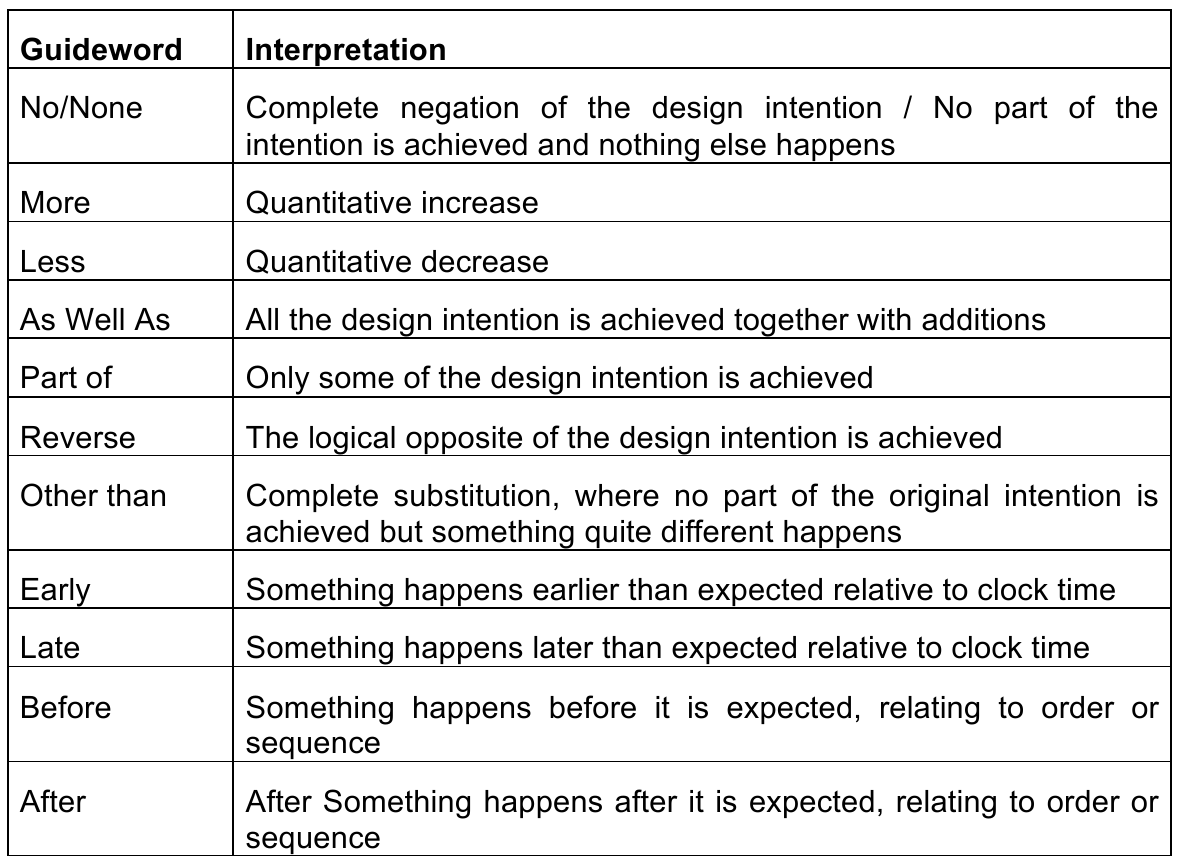}
\caption{Guide words list adapted from \cite{IEC61882}}
\label{fig:guidewordsbase}
\end{figure}

HAZOP (HAZard OPerability) is a collaborative hazard identification technique, developed in the 70's, and is widely used in the process industries. It is now standardized by the standard \cite{IEC61882}. Its success mainly lies in its simplicity and the possibility to apply it at the very beginning of the development process. It is also adaptable to the formalism used to describe a system as presented in the standard \cite{0058}. HAZOP does not consider failure modes as FMECA, but potential deviations of the main parameters of the process. For each part of the system, the identification of the deviation is systematically done with the conjunction of:
\begin{itemize}
\item system parameters, e.g., in the case of an industrial process :  {\tt temperature}, {\tt pressure}, {\tt flow}, etc.
\item guide words like: {\tt No}, {\tt More}, {\tt Less} or {\tt Reverse}
\end{itemize}
The role of the guide word is to stimulate imaginative ideas and initiate discussions. A proposed list of guide words is given in Figure~\ref{fig:guidewordsbase}. For instance, we can have the following conjunctions (e.g., for a chemical process):
\begin{itemize}
\item {\tt Temperature} $\otimes$ {\tt More}  $\rightarrow$ \textbf{\tt Temperature too high}

\item {\tt Flow} $\otimes$ {\tt Reverse} $\rightarrow$ \textbf{\tt Product flow reversal}
\end{itemize}
For each deviation, the procedure is then to investigate causes, consequences and protection, and produce document usually in a table form (similar to FMECA), with columns like: Guide word, Element, Deviation, Possible causes, Consequences, Safeguards, Comments, Actions required, etc.

Even though the HAZOP method has proved to be efficient, the results may be questionable when the boundary of the study is too vast or not well defined, or when the guide words are either too numerous or too limited for the analysis to be relevant. Another limitation is that there is no systematic method to adapt the guide words to the considered domain, so adaptation depends on the expertise of the initiators of the method. Additionally, the HAZOP method needs the allocation of human resources and suffers from combinatorial explosion when too many deviations are considered or when the analysts go into too much details. Hence, the success of a HAZOP study depends greatly on the ability of the analyst and the interactions between team members. The choice of the considered ``system parameters", is of high importance, because all the study relies on it. The HAZOP-UML method proposed in this paper is aimed at providing more guidance to analysts to identify which parameters they have to consider. 

\section{HAZOP-UML}
\label{sec:hazopuml}
One main issue when applying HAZOP is to identify the system parameters. We propose to use UML to partition and describe the system. The considered parameters will be then some elements of the UML diagrams. In this section we will give guidelines to identify those parameters, and the associated guide words to identify possible deviations. This work is the result of several applications and refinement, and may also be completed or modified by the analysts. Even if our objective is to propose a systematic approach, it is important to note that HAZOP-UML does not identify all hazards. First because no single hazard identification technique is actually capable of finding all the hazards \citep{CAN09}, and also because we will focus on the identification of the operational hazards, i.e., hazards linked to the human-robot interactions, through dynamic models of the system. 

As already presented, we propose to focus on the three main dynamic UML diagrams: use case, sequence and state diagrams. For those diagrams, some generic deviations are presented in Section~\ref{sec:gw}. The whole process is then introduced in Section~\ref{sec:process}, and Section~\ref{sec:tool} presents a prototype of a tool for HAZOP-UML.

\subsection{Guide words}
\label{sec:gw}
Instead of using the term ``parameter" usually used in HAZOP studies, entities and attributes of UML elements are introduced in this section. Then for each element, a generic interpretation for a deviation is proposed. This analysis is based on the UML metamodel \citep{OMGUML07}. The selected UML entities are : use case, message, state machine.

\subsubsection{Guide words for use cases}

\begin{figure}
\begin{center}
\includegraphics[width=5cm]{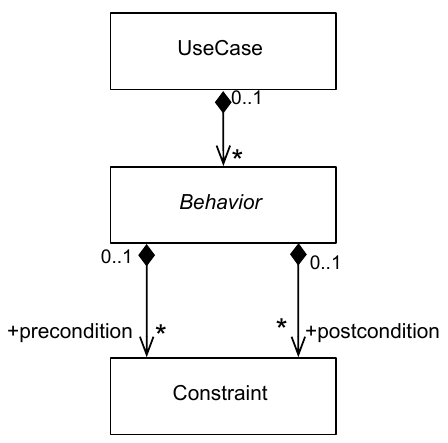}
\caption{Reduced concepts for specification of use cases}
\label{fig:metauc}
\end{center}
\end{figure}
Figure~\ref{fig:metauc} presents an extract from the UML metamodel, focusing on a use case. The UML class diagram notation is used to represent this metamodel. This diagram specifies that a use case may be composed of 0 to several (noted as ``*") \emph{Behaviors}. Indeed, a use case is usually composed of a nominal behavior (or nominal scenario), and several exceptions. Each Behavior may have 0 to several \emph{Constraints}, which are pre and post conditions. As introduced in section~\ref{sec:UML}, we add to this metamodel one constraint to the \emph{Behavior} of a \emph{UseCase}: the invariant. Indeed, when an analyst studies all possible deviations, we would argue that the non-functional requirements, which may be safety invariants (e.g., robot velocity should not exceed 20cm/s) have to be taken into account. We  should then consider that the attributes of a use case are: preconditions, postconditions, and invariants, which are all UML \emph{Constraints}. For this reason, we apply the classical HAZOP guide words to the concept of constraint in a generic way and formulate an interpretation to guide the analyst. The result of this work is given in Table~\ref{fig:gwuc}. 
\begin{table}
\centering
\includegraphics[width=12cm]{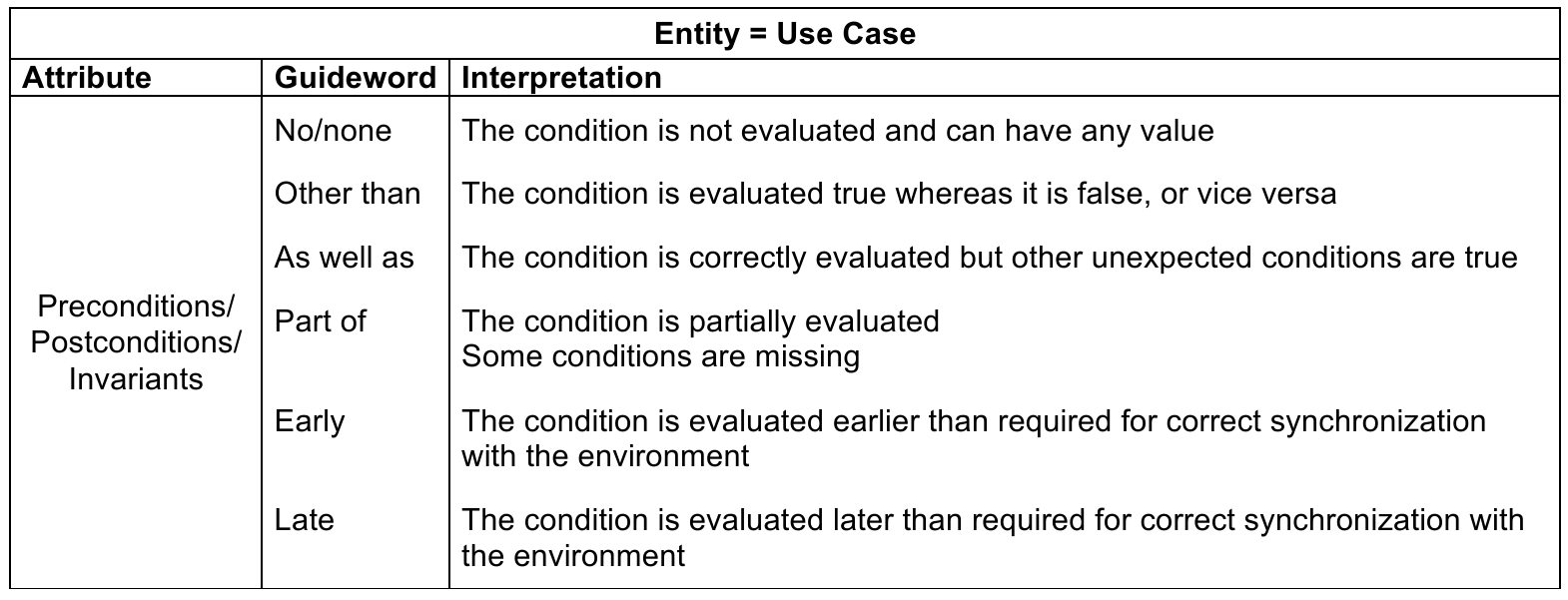}
\caption{Guide words list and generic interpretation for use cases}
\label{fig:gwuc}
\end{table}
Only six guide words were interpreted, we also remove many redundancies in the interpretation. Let consider the example of use case UC02 ("standing up operation") described in Figure~\ref{fig:ucfilehazop}. The precondition ``The robot is in front of the patient" combined with the guide word ``No", leads to the following scenario: the patient tries to standup while the robot is not properly positioned. This might induce excessive effort for the patient and a fall which is catastrophic in our case study. If we consider this use case, with 9 conditions and 6 guide words, this leads to 54 possible deviations. Moreover, the interpretation of a guide word may change from an analyst to another. Nevertheless, the objective is to eventually identify all hazards, and the original guide word used for the identification is of no real importance. 

\subsubsection{Guide words for sequence diagrams}
\begin{figure}
\begin{center}
\includegraphics[width=12cm]{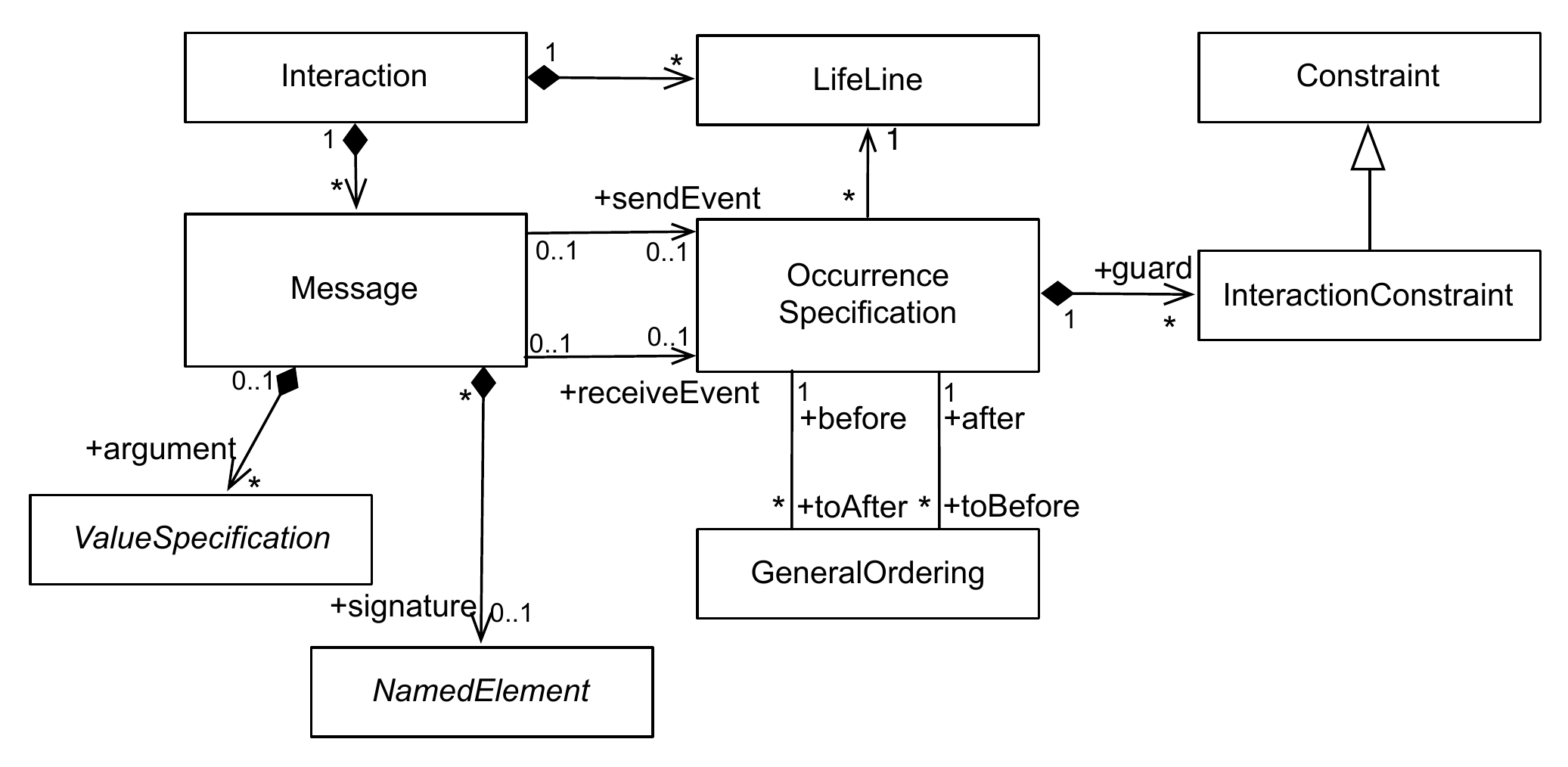}
\caption{Reduced metamodel for interactions in UML (sequence diagrams) extracted from \cite{OMGUML07}}
\label{fig:metasq}
\end{center}
\end{figure}

Sequence diagrams are one of the graphical representation of the \emph{Interaction} UML concept. It is composed of \emph{Lifelines} exchanging \emph{Messages}. This is represented in the simplified metamodel in Figure~\ref{fig:metasq}. This metamodel extracted from \cite{OMGUML07} has very little differences with the version \citep{OMGUML11}, so we kept this representation which is simpler, and expressive enough for its use in HAZOP-UML.
Based on this metamodel, we define five attributes for the Message: 
\begin{enumerate}
\item General Ordering: the general order of the messages within the interaction
\item Send/receive event timing: event related to the clock time
\item Lifelines: send and receiving lifelines of a message
\item Interaction Constraint: guard condition on a message
\item Message argument: parameters of a message
\end{enumerate}
Other elements of the metamodel have not been considered, as we did not find any possible deviation or we intentionally avoid to consider them because they would have produced redundant possible deviations (interested reader may find more about UML \emph{interaction fragments in \cite{OMGUML11}}.
The resulting table for the generic deviations and their interpretation is given in Table~\ref{fig:gwseq}. 
In t\cite{OMGUML11}  the following explanation is given: 
``A \textit{GeneralOrdering} represents a binary relation between two \textit{OccurrenceSpecifications}, to describe that one \textit{OccurrenceSpecification} must occur before the other in a valid trace. 
 This mechanism provides the ability to define partial orders of \textit{OccurrenceSpecifications} that may otherwise not have a specified order."
This could be interpreted as the fact that in some diagrams a \emph{GeneralOrdering} relation can be added as a constraint. But in a sequence diagram, the physical position of the message already specifies an order for a valid trace. 
Hence, in our approach, we will interpret a sequence diagram as a valid trace, i.e., with a valid specified ordering of the message. This trace is descriptive (and not prescriptive like the state machine), but changing the ordering may lead to hazardous interactions.

\begin{table}
\centering
\includegraphics[width=14cm]{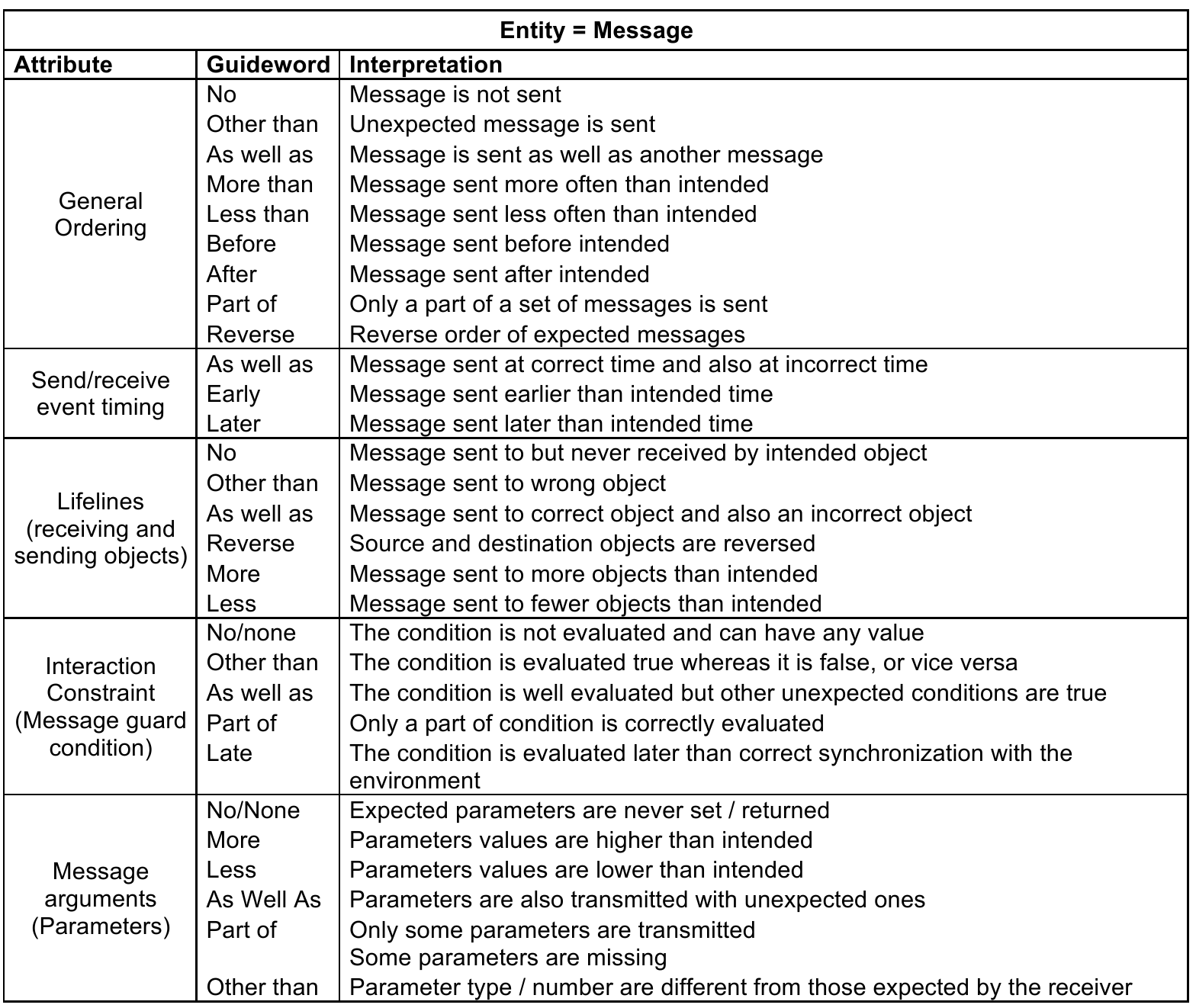}
\caption{Guide words list and generic interpretation for sequence diagram messages}
\label{fig:gwseq}
\end{table}

\subsubsection{Guide words for state machines}
\begin{figure}
\begin{center}
\includegraphics[width=10cm]{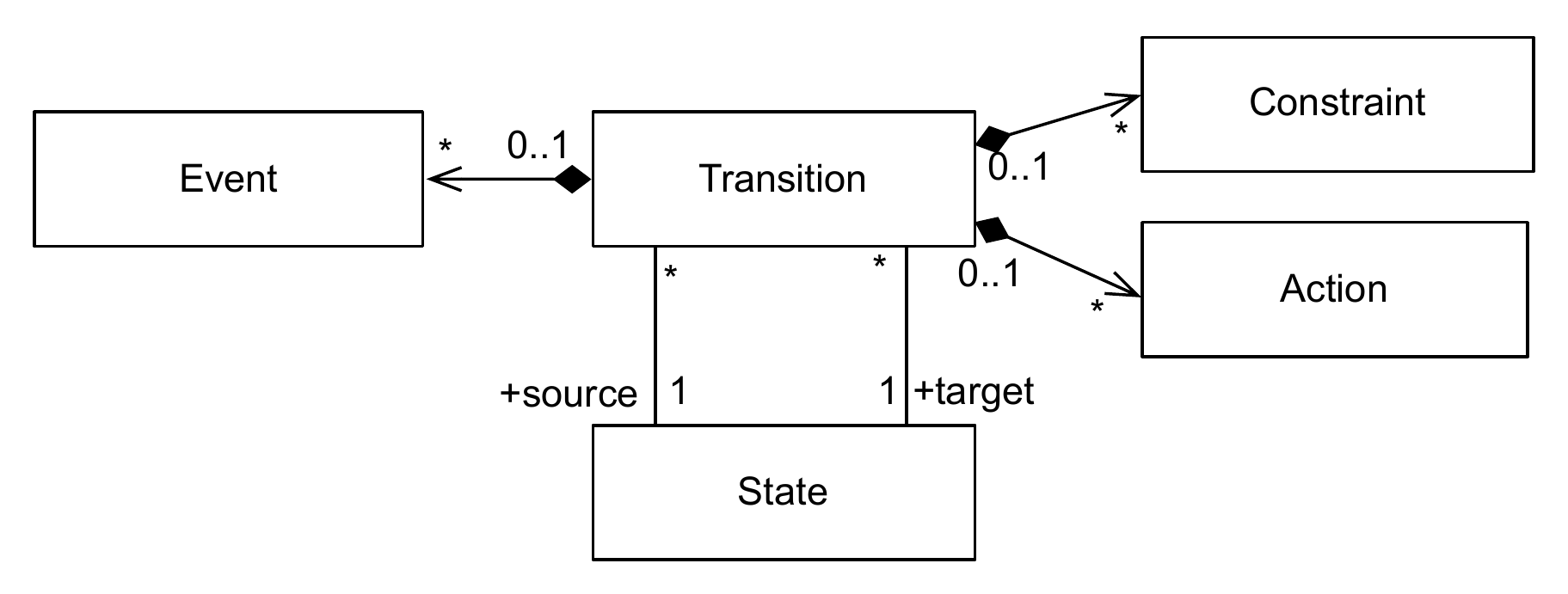}
\caption{Adapted UML metamodel of state machine}
\label{fig:metasm}
\end{center}
\end{figure}

The same approach was used for the state machines. This diagram can also be used for detailed system design, which may lead to a combinatory explosion for the HAZOP analysis. Hence, we reduced the number of concepts to a very simple version as presented in Figure~\ref{fig:metasm}. Note that we replaced in this model the original class \textit{Behavior} by \textit{Action}. Actually, in UML an action is the fundamental unit of behavior specification, which can be associated to a state or a transition. We only consider in this method the action on transitions, which is sufficient to express relevant behavior. Of course, our proposal could be extended to the complete state machine metamodel, to identify all possible deviation at design time, but this is out of the scope of our method.

According to this metamodel, the resulting table for possible deviations is given in Table~\ref{fig:gwstate}.
 \begin{table}
\centering
\includegraphics[width=13cm]{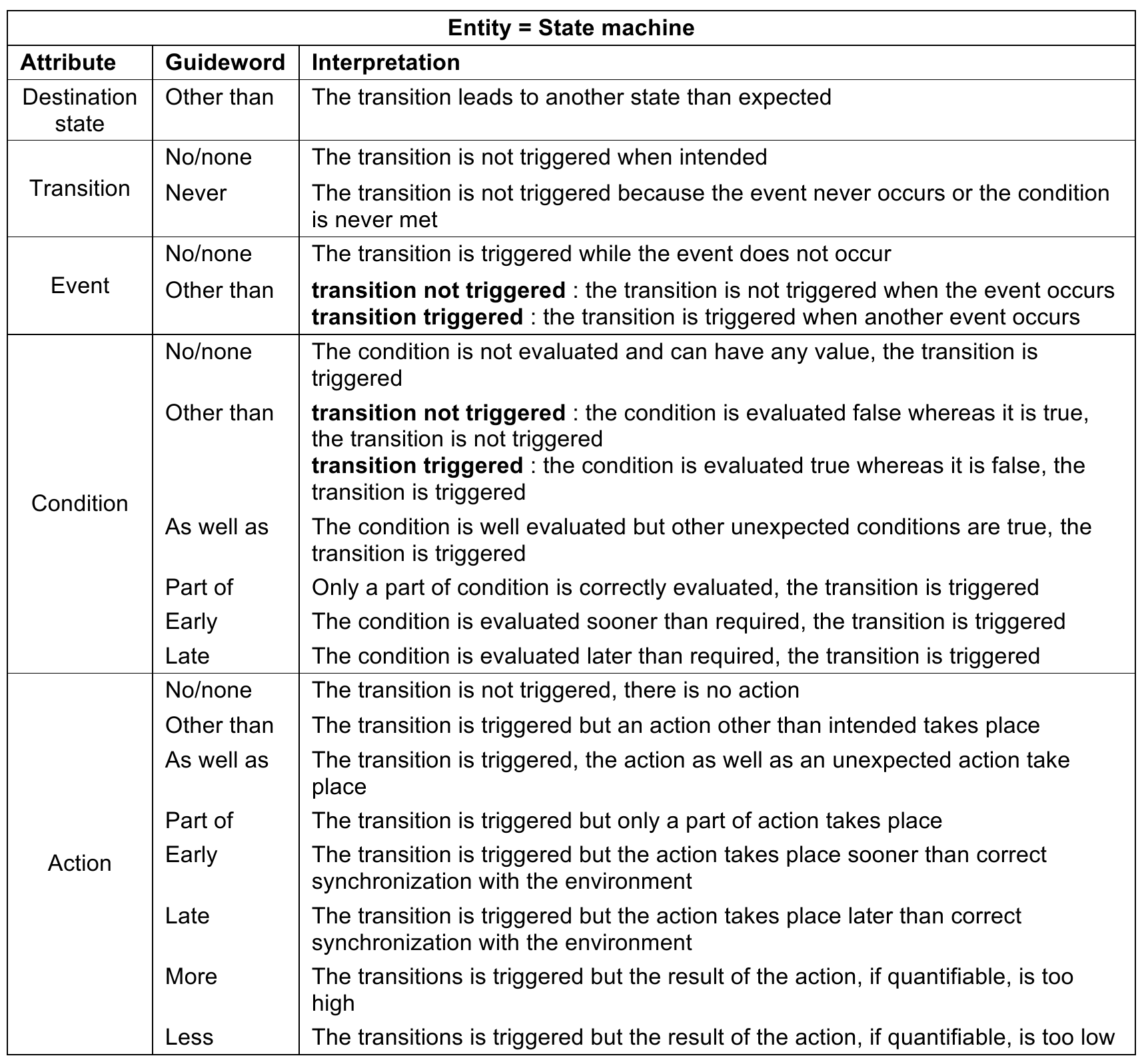}
\caption{Guide words list and generic interpretation for state machines}
\label{fig:gwstate}
\end{table}
In order to provide more guidance, we also point out in this table if the transition is triggered or not for some deviations.

\subsection{HAZOP-UML process and outputs}
\label{sec:process}
\begin{figure}
\centering
\includegraphics[width=14cm]{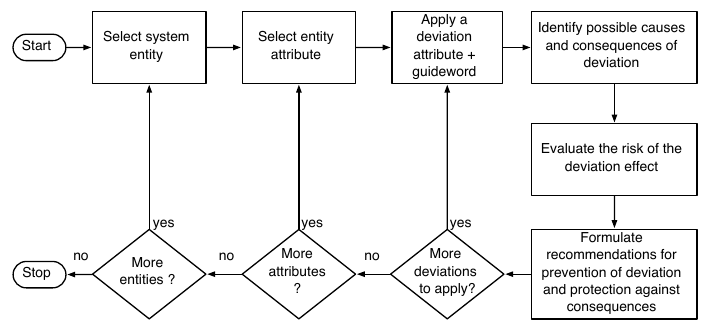}
\caption{HAZOP-UML process}
\label{fig:process}
\end{figure}

\begin{figure}
\centering
\includegraphics[width=\textwidth]{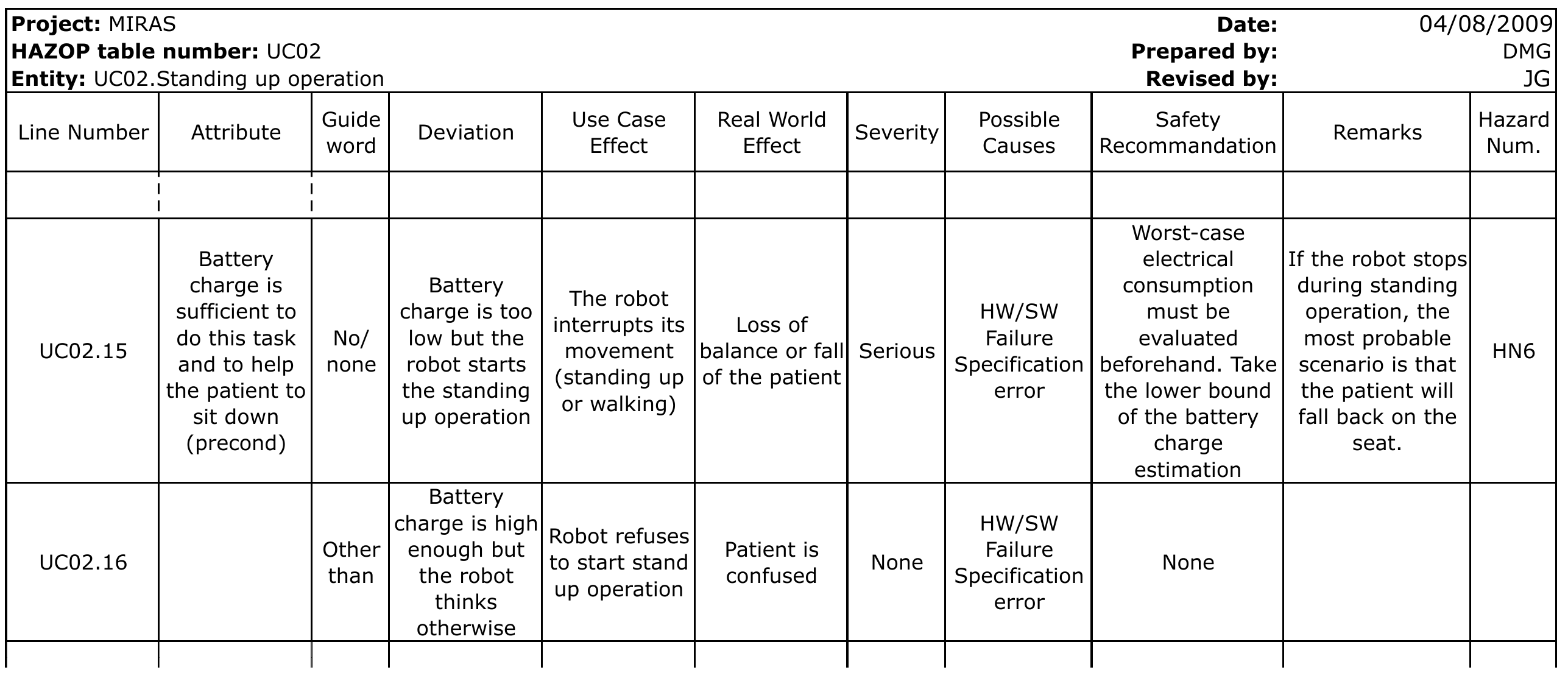}
\caption{HAZOP-UML Table extract}
\label{fig:hazopextract}
\end{figure}

According to the previous tables, the process to perform HAZOP-UML is the following procedure:
for each entity, for each attribute, for each guide words, identify one or several possible deviations and analyse it (them). A graphical view is given in Figure~\ref{fig:process}.
The analysis of the deviation may include the identification of possible causes and consequences. Depending on the project, it is also possible to evaluate the risk (consequence of the deviation effect, and likelihood of the considered deviation). Nevertheless, this information is usually too complex or impossible to obtain. On the contrary, such analysis always includes identification of recommendations to treat the deviation or its causes or it consequences (prevention and protection means). To establish such a study, the columns of a table as in Figure~\ref{fig:hazopextract} are given hereafter:
\begin{enumerate}
\item Entity: the UML element on which the deviation is applied (here UC02 is the same for all the table so it is in the head of the table) 
\item Line number: for traceability (UCx.line\_number)
\item Attribute: the considered attribute (e.g., a use case precondition)
\item Guide word: the applied guide word
\item Deviation: the deviation resulting from the combination of the entity attribute and the guide word based on Tables~\ref{fig:gwuc}, \ref{fig:gwseq} and \ref{fig:gwstate}.
\item Use Case Effect: effect at the use case level.
\item Real World Effect: possible effect in the real world.
\item Severity: rating of effect of the worst case scenario in the real world.
\item Possible Causes: possible causes of the deviation (software, hardware, human, etc.).
\item Safety Recommendations for prevention or protection
\item Remarks: explanation of analysis, additional recommendations, etc.
\item Hazard Numbers: real world effects are identified as hazards and assigned a number,
helping the users to navigate between results of the study and the HAZOP-UML tables.
\end{enumerate}
In Figure~\ref{fig:hazopextract} given example, a precondition of UC02 (previously presented in Figure~\ref{fig:ucfilehazop}) is analyzed using the guide words No and Other than. It leads to identify the hazard HN6 (Fall of the patient due to imbalance caused by the robot).

\begin{figure}
\centering
\includegraphics[width=10cm]{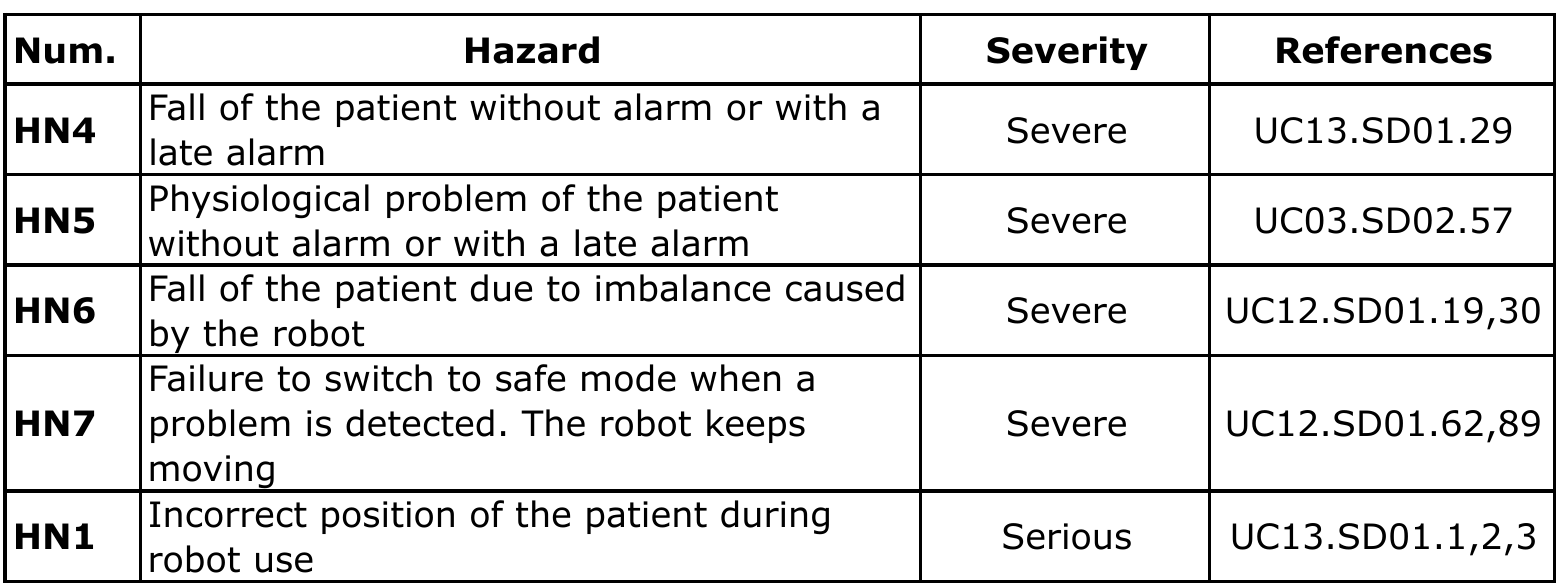}
\caption{Hazard list extract}
\label{fig:hazardlist}
\end{figure}

\begin{figure}
\centering
\includegraphics[width=10cm]{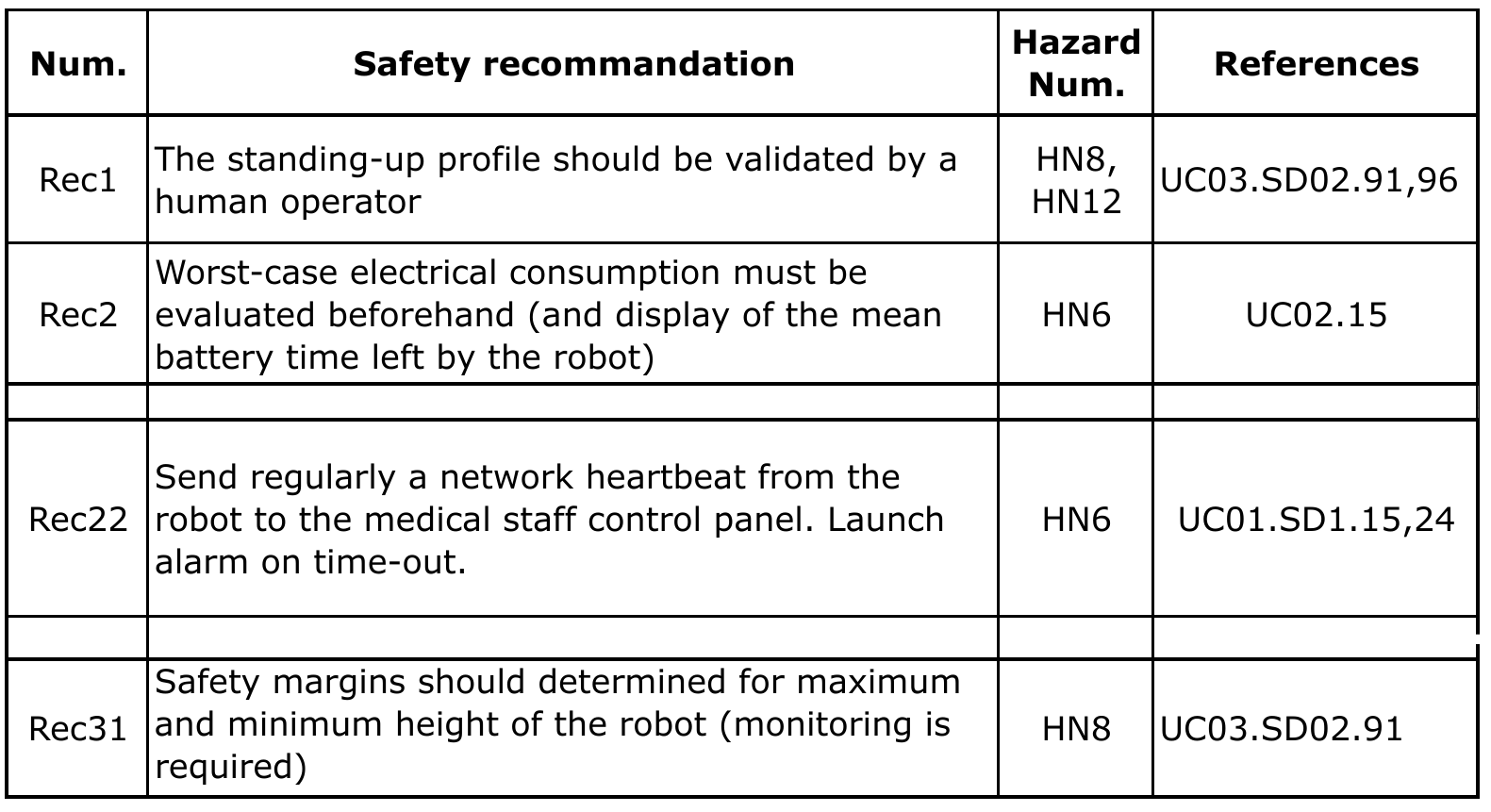}
\caption{Recommendation list extract}
\label{fig:reco}
\end{figure}

The resulting documents are the tables as the raw artefacts, but also:
\begin{itemize}
\item a concatenated list of identified hazards
\item a list of hypotheses made to perform the analysis, which need to be confirmed by domain experts to validate the study
\item a list of safety recommendations
\end{itemize}
All those documents reference each others using numbered labels for lines, hazards (HN), recommendations (Rec), and hypothesis. Examples of a hazard table and recommendation list are given in Figure~\ref{fig:hazardlist} and Figure~\ref{fig:reco}. As an example, recommendation Rec2 from Figure~\ref{fig:reco}, covers hazards HN6 (fall of the patient), and has been formulated in the HAZOP table UC02 line 15 (UC02.15).

\subsection{A tool for HAZOP-UML}
\label{sec:tool}
\begin{figure}
\centering
\includegraphics[width=18cm,angle=90]{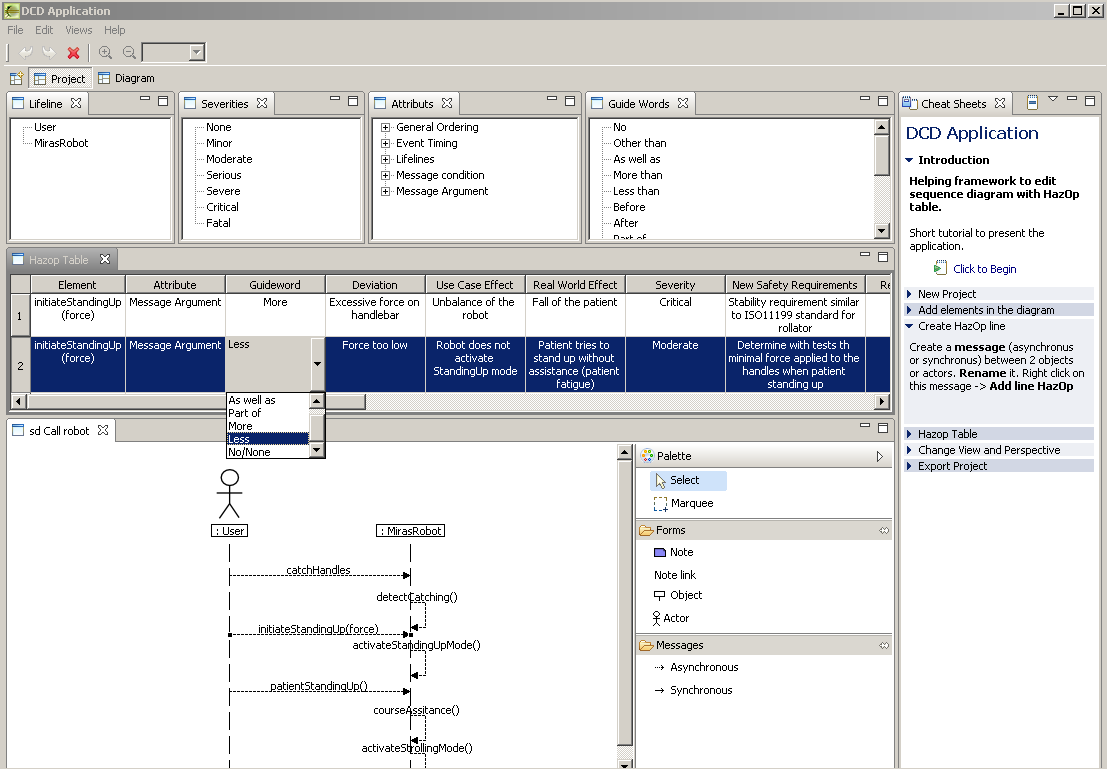}
 \caption{Main view of the tool to support the HAZOP-UML method}
\label{fig:tool-general}
\end{figure}

To ease the analysis of complex systems, we developed a prototype of a tool to
support the method. It helps to manage the
combinatorial aspects of the HAZOP method by maintaining consistency
between UML models and HAZOP tables and by providing document
generation and management features. The tool is built as an
Eclipse plugin (www.eclipse.org) using the Graphical Modelling
Framework (GMF). In this tool presented in Figure~\ref{fig:tool-general}, the analyst can draw
UML use case and sequence diagrams. Using guide word templates, HAZOP
tables are automatically generated, ready to be filled out by the analyst using choice lists.

The list of guide words, the list of columns and the list of severities are editable using the main
project view. Using the
template, the analyst can add a line in the table by selecting a message, and then select applicable
deviations and fill in the corresponding columns. 
When completing the table, the recommendation list
and corresponding hazards are automatically generated in
the project view. The toolbox of the HAZOP guide words allows
deviations to be added (for example, several deviations for the same
keyword). Finally a report in HTML can be generated consisting of HAZOP tables, UML diagrams, and hazards, recommendations and hypotheses lists.

\section{Experiments and results}
\label{sec:exp}
This section provides results of the experimentation of HAZOP-UML on three robotic applications developed within the following projects:
\begin{itemize}
\item ANR-MIRAS (Multimodal Interactive Robot of Assistance in Strolling) \citep{MIRAS} an assistive robot for standing up, sitting down and strolling already presented in Section~\ref{sec:UML}.
\item FP6-PHRIENDS (Physical Human-Robot Interaction: depENDability and Safety) \citep{PHRIENDS}. The system is a mobile robot with a manipulator arm. The considered environments are workshops and factories with human workers. Collaborative work between a human and a robot is possible (e.g., the robot can give an object to the human). The arm is the KUKA Light Weight Robot (LWR), a seven degrees of freedom arm which contains torque and motor position sensors. The mobile base is the KUKA omnirob product. 
\item  FP7-SAPHARI (Safe and Autonomous Physical Human-Aware Robot Interaction) \citep{SAPHARI}. As in PHRIENDS, an Industrial co‐worker operates in a manufacturing setting accessible to human workers. The mobile manipulator may encounter humans while moving between the different workstations because the operation area is freely accessible to human workers. 
It takes and places part boxes on shelves, work stations, or on the robot base in order to convey them. The robot navigates autonomously in its operation area. When the robot encounters unexpected or difficult situations the worker might intervene and help by giving the robot direct haptic instructions. 
\end{itemize}

For all three experiments, we followed the same procedure. We recruited analysts (an engineer for PHRIENDS, a postdoctoral for MIRAS, and a Phd student for SAPHARI), who were trained in our laboratory to HAZOP-UML. As a first step, they were in charge of modeling the UML diagrams, and validate them with robotic and domain experts (for instance in MIRAS, validation was also performed by doctors from the hospitals of the project). A second step was the deviation analysis performed only by the recruited analyst, followed by a revision by another member of our laboratory already trained to HAZOP-UML. Then, the resulting hazard and recommendation lists were discussed and validated by the robotic and domain experts. Quantitative data (e.g., working time or numbers of deviations) and qualitative data (e.g., traceability or modifiability) coming from these experiments are presented in this section, and structured according to the following properties:
\begin{itemize}
\item Applicability: we estimated the resources needed for the application of HAZOP-UML
\item Guide words relevance: this is a critical point of the method as all the results will depend on the ability of those guide words to guide the analyst
\item Validity: we compared results from a Preliminary Hazard Analysis to HAZOP-UML to assess its validity. 
\item Usability: some benefits and limits of HAZOP-UML while using it.
\end{itemize}

\subsection{HAZOP-UML applicability}

Classic HAZOP is usually applied in collaborative workshops, involving many partners to maximize the chances of study completeness. On the contrary, HAZOP-UML can be applied by a single analyst and then validated by experts. This comes from the fact that the study is always based on a UML model, which has been done in collaboration with stakeholders (e.g., robotic engineers or medical staff). The fact that their knowledge has been captured by UML models, makes the safety analyst task more independent from domain experts. Of course, during the analysis several questions arise, and hypotheses need to be made to carry out the analysis. They need then to be validated by the experts (this is why we propose to produce a hypotheses list). 

Considering that a single analyst can perform most of the work, we also evaluate the effort to perform the complete analysis. Numbers are given in Table~\ref{tab:appli-stats} for the three robotic projects. The state-machine version of HAZOP-UML has only been applied to MIRAS and statistics are presented in Table~\ref{tab:appli-stats-etats}.

\begin{table}
  \centering
\scriptsize 
\usefont{T1}{phv}{l}{n}
\begin{tabular}{|l|c|c|c|} 
\cline{2-4}
\multicolumn{1}{c|}{}                                       & PHRIENDS & MIRAS  & SAPHARI\\ \hline \hline
\textbf{Use cases}                                    & 9              & 11    &  15   \\ \hline
Conditions                                   & 39            &  45     &  54 \\ \hline
Analyzed deviations                  & 297          &  317     & 324 \\ \hline
Interpreted deviations &  179  & 134  &  65  \\ \hline
Interpreted deviations with  &  120   &  72  & 50\\
recommendation & & &\\ \hline \hline
\textbf{Sequence diagrams}                   &  9              &  12   &    16  \\ \hline
Messages                                    & 91            &  52   &    122  \\ \hline
Analyzed deviations                    & 1397        &  676    &    2196    \\ \hline
Interpreted deviations   & 589   &  163 & 87 \\ \hline
Interpreted deviations with    &  274 &  85  & 36 \\ 
recommendation & &  &\\ \hline \hline
Number of hazards & 21 & 16 & 28\\ \hline
\end{tabular}
\normalsize
\normalfont
 \caption{Statistics for the application of HAZOP-UML for the three projects}
\label{tab:appli-stats}
\end{table}

\begin{table}
  \centering
\scriptsize 
\usefont{T1}{phv}{l}{n}
\begin{tabular}{|l|c|} 
\cline{2-2}
\multicolumn{1}{c|}{} & MIRAS \\ \hline \hline
State Machine diagram      & 1                      \\ \hline
States                                   & 9                \\ \hline
Transitions                                   & 19      \\ \hline
Analyzed deviations                  & 215   \\ \hline
Interpreted deviations with  &  161 \\
recommandation &  \\ \hline
\end{tabular}
\normalsize
\normalfont
 \caption{Statistics for the application of HAZOP-UML State-machine only to MIRAS}
\label{tab:appli-stats-etats}
\end{table}
For the three projects, the complexity was nearly the same (between 39 and 54 use case conditions, and 91 and 122 messages in sequence diagrams). For each project one analyst has been recruited. Those three analysts were a post-doctoral, an engineer, and a Dr-engineer. ``Analyzed deviations" stands for the number of deviations the analyst has considered, but only a part of them leads to an `Interpreted deviations". 

The resulting numbers show that no combinatory explosion happened, and less than 0.5 man-month was necessary for each study. Few iterations for table updates were needed (between 2 and 3). The presented tool in Section~\ref{sec:tool} was under development during those three projects, so we used a classic spreadsheet software with templates and macros. The cross checking between HAZOP tables and UML diagrams was then done by hand, which is clearly a limit that we want to reduce with our tool. Same conclusions were drawn for the state machine study, which was only applied to the MIRAS project (Table~\ref{tab:appli-stats-etats}). However, those three projects were successful regarding the applicability of our method.

\subsection{HAZOP-UML guide words relevance}
For all projects, statistics of guide word usage have been made. The results of PHRIENDS project are presented in Tables~\ref{fig:phriendsseqgw} and \ref{fig:phriendsucgw}. A first remark is that most of the guide words have been used by the analyst except in some special cases. The lifeline attribute is particularly useful when the 
robotic system is communicating with different actors (e.g., other robots), which was not the case in our project. The PHRIENDS UML diagrams also did not include any constraint on the messages, so the ``Interaction constraints" guide words weren't used either in our case study. The guide word ``Less than" (Message sent less often than intended) was also not used, as no constraint on frequency for messages was specified in the UML diagrams. The analyst also considered that ``Part of" (only a part of a set of message is sent) was not relevant, because the level of description of UML diagram did not allow to consider parts of a message (as it may be the case with complex message sending with long protocol). Nevertheless, we chose to keep these guide words as in some special cases they would be applicable. 

Another result, which is not presented here, is the redundancy of the hazards found, with different guide words. This is actually not an issue, because our main objective is to find a list of hazards, whatever guide word used to identify it. To determine if the guide words list is not limiting, we only rely on the results of the application on the three projects. A formal demonstration is actually impossible, and as already discussed, no single hazard identification technique is actually capable of finding all the hazards. We thus consider that in order to propose a systematic approach, the selected guide words are sufficient to identify all the major hazards.

\begin{table}
\centering
\includegraphics[width=8cm]{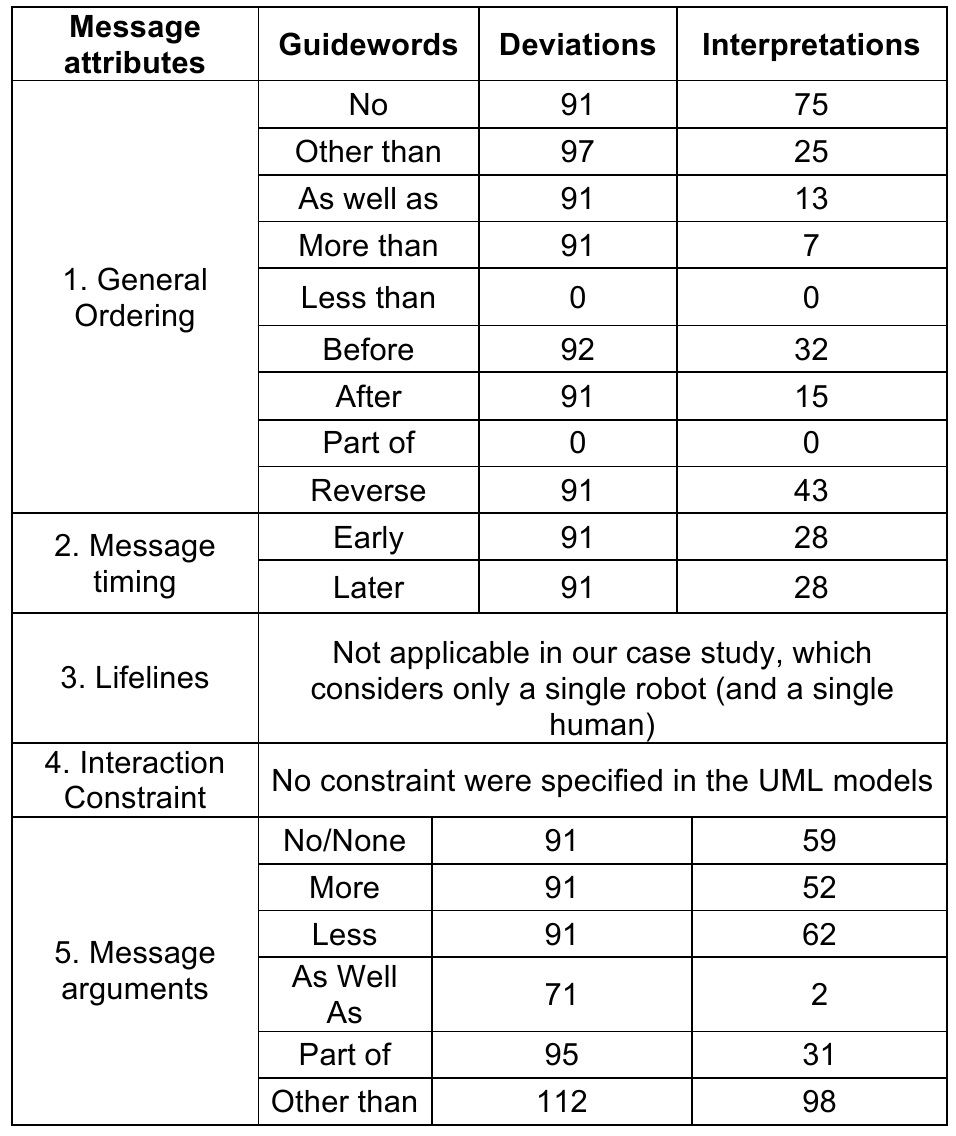}
\caption{Sequence diagram guide words utility in PHRIENDS}
\label{fig:phriendsseqgw}
\end{table}

\begin{table}
\centering
\includegraphics[width=9cm]{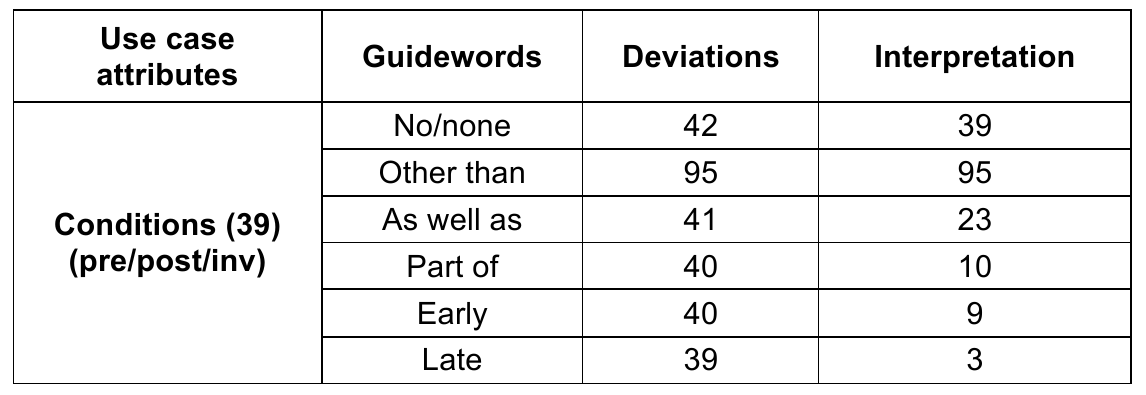}
\caption{Use case guide words utility in PHRIENDS}
\label{fig:phriendsucgw}
\end{table}

\subsection{HAZOP-UML validity}
\label{sec:validity}
Table~\ref{fig:hazardslist} presents two results for validity. First, this study shows that all hazards found during the PHA (Preliminary Hazard Analysis), done by collaborative workshop between a safety analyst and robotic experts, were also identified during HAZOP-UML (performed by the analyst), and that new hazards were also found. The fact that all scenarios of use were modeled in UML significantly improves the analysis. For instance, the hazard HN11 (Disturbance of medical staff during an intervention), was only identified during use case analysis, and never mentioned during the PHA, whereas it is highly relevant in case of emergency intervention. 

The second analysis presented in this Table shows that use cases (UC) and messages (Seq) analysis are complementary, whereas state machine analysis has a redundant contribution for hazard identification. For instance, HN4 identified 11 and 13 times during use case and sequence diagrams analyses, has been identified 32 more times during state machine analysis.   Nevertheless, we believe that state machine analysis is also interesting to identify more sources of deviations that could be used in other risk analysis methods, and also provide safety recommendations which are different from use cases and messages ones.

\begin{table}
\centering
\includegraphics[width=11cm]{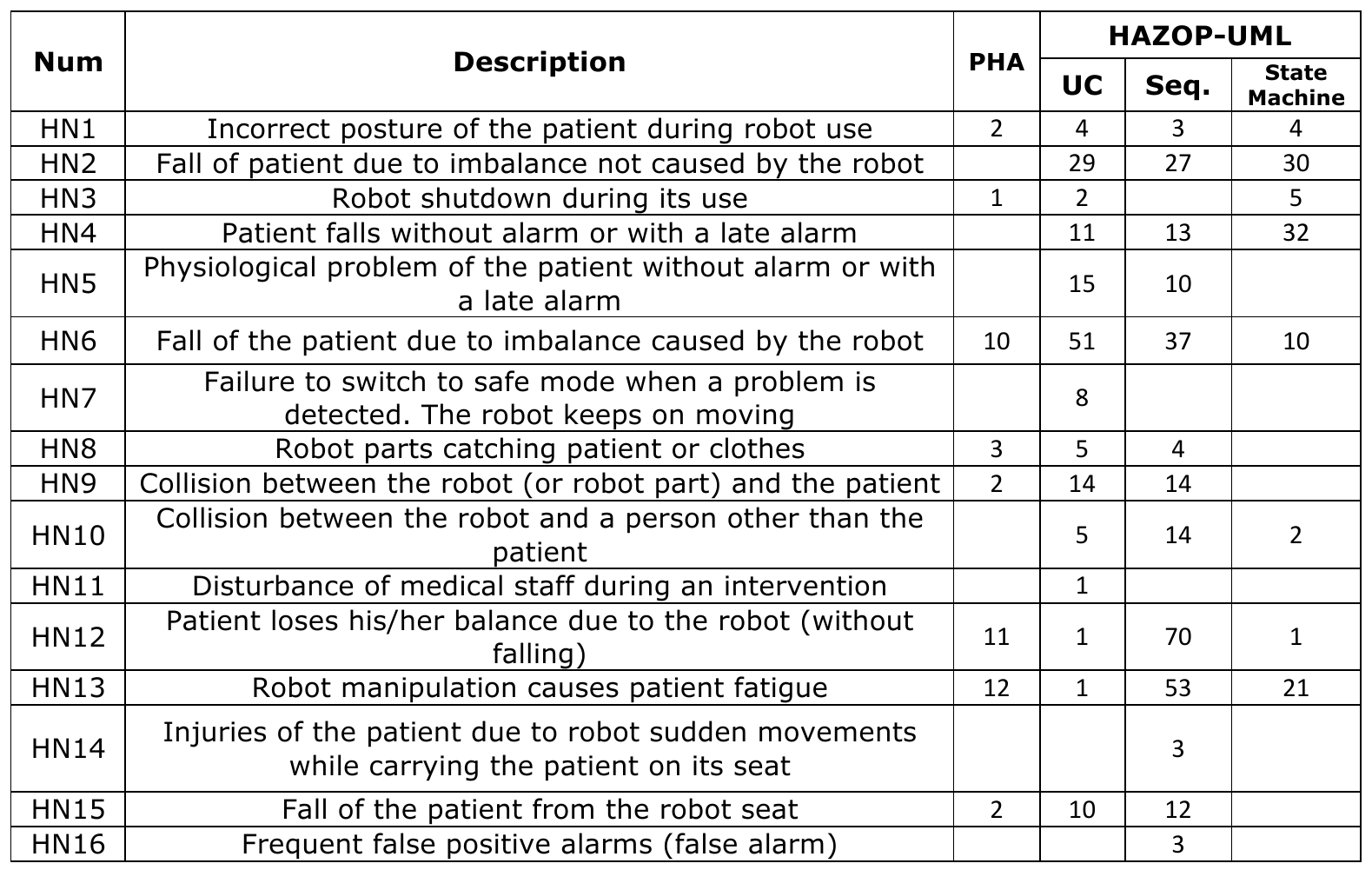}
\caption{Hazard list and occurrences in PHA and HAZOP-UML in MIRAS}
\label{fig:hazardslist}
\end{table}

\subsection{HAZOP-UML usability}

A major advantage of HAZOP-UML lies in its simplicity. Indeed, UML models have been simplified to be easily understandable by non experts without reducing its expressiveness. HAZOP is also an intuitive method. Several engineers from different domains (electronics, computer science or risk management) have been trained to the method in few days. 

HAZOP-UML is completely integrated and consistent with the development process. Indeed, same UML diagrams were used in the projects, to define the scenarios. This helped us for each iteration in the development process to easily update the HAZOP tables. This traceability is an important issue in safety analysis methods, which are usually applied once due to the cost to apply them. 

Among HAZOP-UML limitations, we remind that HAZOP-UML is focusing on operational hazards (linked with the robot tasks). We thus do not consider ``machine" hazards already defined in many standards, like electrocution, explosion, etc. As already mentioned, this method should be completed by other hazard analysis techniques. A second limitation is the fact that the UML models and HAZOP tables do not explicitly mention the environment conditions of execution. For instance, a similar scenario but with high or low level of light might change the deviations and their consequences. It is still an open issue and an integration in the UML models would be an interesting direction. Last but not least, the HAZOP-UML has the same drawback as other risk analysis methods, which is a difficult determination and expression of the hazard because of the fuzziness of a hazard definition (``potential source of harm", from \cite{Guide51ang}) which may designate both a cause or a consequence. Three columns in the HAZOP table can represent a hazard: deviation, use case effect, real word effect. In many tables, we found that some real word effects were already mentioned as use case effects in other HAZOP table lines. We chose to reduce the number of hazards, taking into account only the ``real word effect" as a hazard, but for some cases where it was obvious that the treatment would be completely different, we also took into account the deviation and use case effect. For instance, 
in Table~\ref{fig:hazardslist}, the hazard HN2 (Fall of patient due to imbalance not caused by the robot) and HN6 (Fall of the patient due to imbalance caused by the robot), lead both to the fall of the patient, but have been differentiated. Even if we provide a well guided method, extraction and formulation of hazards list require a high level of expertise from the safety analyst, in order to choose the right level of description of a hazard.

\section{Related work on model-based hazard identification, tools and methods}
\label{sec:relatedwork}

This section presents related work, focusing on model-based safety analysis, and more particularly those using UML. 
The concept of ``model-based" refers to the fact that a safety analysis technique (e.g., FTA) is based on an abstract representation of the studied system. This was already done at the very first hours of the risk analysis techniques using for instance block diagrams, or had-hoc representations. The quite recent model-based term, usually refers to the use of standardized models (like UML) and the possibility to have tools assisting analysts to produce automatic, or semi-automatic safety analysis based on a system model. Generally, model-based safety analyses focus on the following issues \citep{BLA10}:
\begin{enumerate}
\item Fault propagation analysis
	\begin{enumerate}
		\item \emph{bottom-up}: a fault effect on the system
		\item \emph{top-down}: induction of faults inducing an unwanted effect
	\end{enumerate}
\item Dependability (or safety) properties verification
\item Quantification of probability of unwanted events
\end{enumerate}

Many high-level modeling languages for safety analyses have been defined to cover those points. Just to cite some of them, HIPS-HOPS (\emph{Hierarchically Performed Hazard Origin and Propagation Studies}) and its associated tool developed at Hull university \footnote{\url{http://hip-hops.eu} (accessed 2015-05-15)}, automatically generates fault trees and FMECA tables starting from system models (e.g., Simulink models). For each component, fault annotations are given, and the tool propagates those faults to build safety models (e.g., Fault trees). Altarica \citep{BOI06, LIP15} provides means for fault tree generation or properties verification from system and reliability models. Additionally, many European research projects addressed model-based safety analysis: ESACS (2001-2003)\footnote{\url{www.transport-research.info/web/projects/project_details.cfm?ID=2658}} in transportation domain, followed by ISAAC (2004-2007) \footnote{\url{http://ec.europa.eu/research/transport/projects/items/isaac_en.htm}} in avionics, then CESAR (2009-2012) \footnote{\url{www.cesarproject.eu}} followed by CRYSTAL (2013-2017) \footnote{\url{www.crystal-artemis.eu}} for embedded systems. Previous techniques and works, usually rely on a precise description of the system behavior, which is usually not available at the beginning of a human-robot project.

The method put forward in this paper falls within the scope of fault propagation analysis, and can be described as a ``middle-up approach", as we do not start from ``faults" but from deviations. Our objective is then to identify hazards (and hazardous situations) during human-robot interaction. A very close work is advanced by \cite{LEV11}, with a method called STPA (System Theoretic Process Analysis), which provides guidance to users combining guide words (like in HAZOP) and fault models, applied to models, based on a process/controller/actuator/sensor representation. Many recent applications of STPA can be found, e.g., in robotics \citep{ALE15}, space \citep{ISH10}, railway \citep{THO11} or automotive \citep{SUL14}. One difference with our approach is that scenarios are actually not modeled in this approach. Users are represented as ``controllers", which is not clear while describing human-robot interactions. STPA objective is also different in the way that it really focuses on the identification of cause-consequence chain, which is not the objective of HAZOP-UML (only find the hazards and hazardous situations). We also propose to use UML which is not the case in STPA. On the contrary, the work done in the CORAS project \citep{CORAS, GRA04}, is based on UML to analyse security. Even if we focus on safety, our objectives are the same. A major difference is that we strongly interconnect UML
models and the risk analysis technique HAZOP, which was not addressed in CORAS. 

Our risk analysis  approach is based on a re-interpretation of HAZOP guidewords in the context of some 
UML diagrams. A similar approach has been followed in some previous studies considering UML structural diagrams \citep{HAN04, GOR05,GOR06} and dynamic diagrams \citep{JOH01,ALL01,ARL06,IWU07,SRI05}. In all those papers, the guide words were quite reduced (e.g., only omission and commission) or the link with UML language elements was not fully explored. 
We actually extended the results of those studies, focusing only on use case, sequence and state machine diagrams, in order to explore deviations during operational life. We also paid a particular attention to the human errors expression and analysis in this method, which was absent from the previous papers.

\section{Conclusion}
\label{sec:conclusion}

We set forth a new method for the safety analysis of human-robot interaction called HAZOP-UML. To build this method we used the UML metamodel to identify the basic elements of three dynamic models. We then proposed three guide words tables for use cases, messages of sequence diagrams, and state machines. Those guide words tables help the safety analyst to imagine possible deviations for every elements of those dynamic models. Those deviations are then reported in HAZOP tables, where causes, consequences, and recommendations are formulated. This process produces lists of hazards,  recommendations, and hypotheses. 

This method has been applied successfully on several projects, and we present in this paper a general analysis of the benefits and the limits of the method. We particularly focus on the applicability and validity of the approach. Main advantages of HAZOP-UML are:
\begin{itemize}
\item simple (training and application)
\item applicable at the first step of the development process
\item limits the combinatory explosion
\item consistent with system models, and inherits of system modeling benefits: traceability and modifiability
\item easily supported by a computer assisting tool
\end{itemize}
Even if the models and HAZOP tables can be easily achieved, the main limit lies in the necessity of a high expertise to formulate hazards from HAZOP tables. It is up to the safety analyst to determine the right level of detail for the hazard identification. 

Additionally to the three projects presented in this paper, HAZOP-UML has also been used as a first step of a method to build independent safety monitors in the context of autonomous robots \citep{MAC14}, and we also plan to use it as an entry point for defining virtual words for testing mobile robots in simulation. A future direction is the complete transfer to industry, which is already started in the project \cite{CPSLab}.

\section*{Acknowledgments}
This work was partially supported by the MIRAS project, funded under ANR-TecSan 2009 framework, and the PHRIENDS and SAPHARI Projects, funded under the 6th and the 7th Framework Programme of the European Community.

\section*{References}

\bibliographystyle{elsarticle-harv} 
\bibliography{bib}

\end{document}